\documentclass[letterpaper]{article}
\usepackage{aaai}
\usepackage{graphicx}
\usepackage{times}
\usepackage{helvet}
\usepackage{courier}
\usepackage{amsmath} 
\usepackage{amssymb}  
\usepackage{amsthm} 
\usepackage{alltt}
\usepackage{xspace}
\usepackage{url}
\usepackage{html}
\usepackage[table]{xcolor}
\usepackage[colorlinks=true,citecolor=blue,linkcolor=blue,urlcolor=blue]{hyperref}
\usepackage{epstopdf}

\newcommand{\blue}[1]
           {\color{black} #1 \color{black}}

\newcommand{\myquote}[1]{\vskip6pt\noindent\hangindent=1.2em\hangafter=0\parbox{3.1in}{#1}\vskip6pt}

\nocopyright

\begin{document}


\title{\emph{Herding the Crowd}: Automated Planning for Crowdsourced
Planning}


\author{Kartik Talamadupula \and Subbarao Kambhampati \\
 {\bf Dept. of Computer Science and Engg.}\\
 Arizona State University\\
 Tempe, AZ 85287 USA\\
 \begin{normalsize}{\tt \{krt, rao\}} @ {\tt asu.edu}\end{normalsize}}


\maketitle


\begin{abstract}

  There has been significant interest in crowdsourcing and human
  computation. One subclass of human computation applications are
  those directed at tasks that involve planning (e.g. travel planning)
  and scheduling (e.g. conference scheduling). Much of this work
  appears outside the traditional automated planning forums, and at
  the outset it is not clear whether automated planning has much of a
  role to play in these ``human computation'' systems.  Interestingly
  however, work on these systems shows that even primitive forms of
  automated oversight of the human planner does help in significantly
  improving the effectiveness of the humans/crowd.  In this paper, we
  will argue that the automated oversight used in these systems can be
  viewed as a primitive automated planner, and that there are several
  opportunities for more sophisticated automated planning in
  effectively steering crowdsourced planning.  Straightforward
  adaptation of current planning technology is however hampered by the
  mismatch between the capabilities of human workers and automated
  planners. We identify two important challenges that need to be
  overcome before such adaptation of planning technology can occur:
  (i) {\em interpreting} the inputs of the human workers (and the
  requester) and (ii) {\em steering} or critiquing the plans being
  produced by the human workers armed only with {\em incomplete}
  domain and preference models. In this paper, we discuss approaches
  for handling these challenges, and characterize existing human
  computation systems in terms of the specific choices they make in
  handling these challenges.

%
\end{abstract}



\section{Introduction}
\label{sec:intro}


In recent years, thanks to the ease of communication afforded by the
internet, human computation has emerged as a powerful and inexpensive
alternative to solving computationally hard problems, especially those
that require input from humans for solution. Indeed, the area has been
defined as {\em `` ...a paradigm for utilizing human processing power
  to solve problems that computers cannot yet
  solve.''}~\cite{von2009human}. A similar term, crowdsourcing, is
often used to denote the process wherein traditional (perhaps
specifically trained or skilled) human workers are replaced by members
of the public~\cite{howe2006crowdsourcing}. To complete the
classification~\cite{quinn2011human}, collective intelligence is a
label that is often given to a set of tasks which contain {\em
  ``...groups of individuals doing things collectively that seem
  intelligent''}~\cite{malone2009harnessing}. A core class of human
computation problems are thus directed at that quintessential human
activity: {\em planning}.
Several recent efforts have started looking at crowd-sourced planning
tasks
\cite{law2011towards,zhang2012human,cobi,lasecki2012real,lotosh2013crowdplanr}.




At first glance, these applications and problem categories appear to
have very little to do with existing automated planning methods, as
they seem to depend solely on human planners. However, a deeper look
at these systems shows that most of them use primitive automated
components in order to enforce checks and constraints to steer human
workers.  More importantly, experiments show that even these primitive
automated components go a long way towards improving plan quality, for
little to no investment in terms of cost and time
(c.f. Zhang et al.~\citeyear{zhang2012human}).





\begin{figure*}[t]
  \begin{center}
    \includegraphics[width=7.0in]{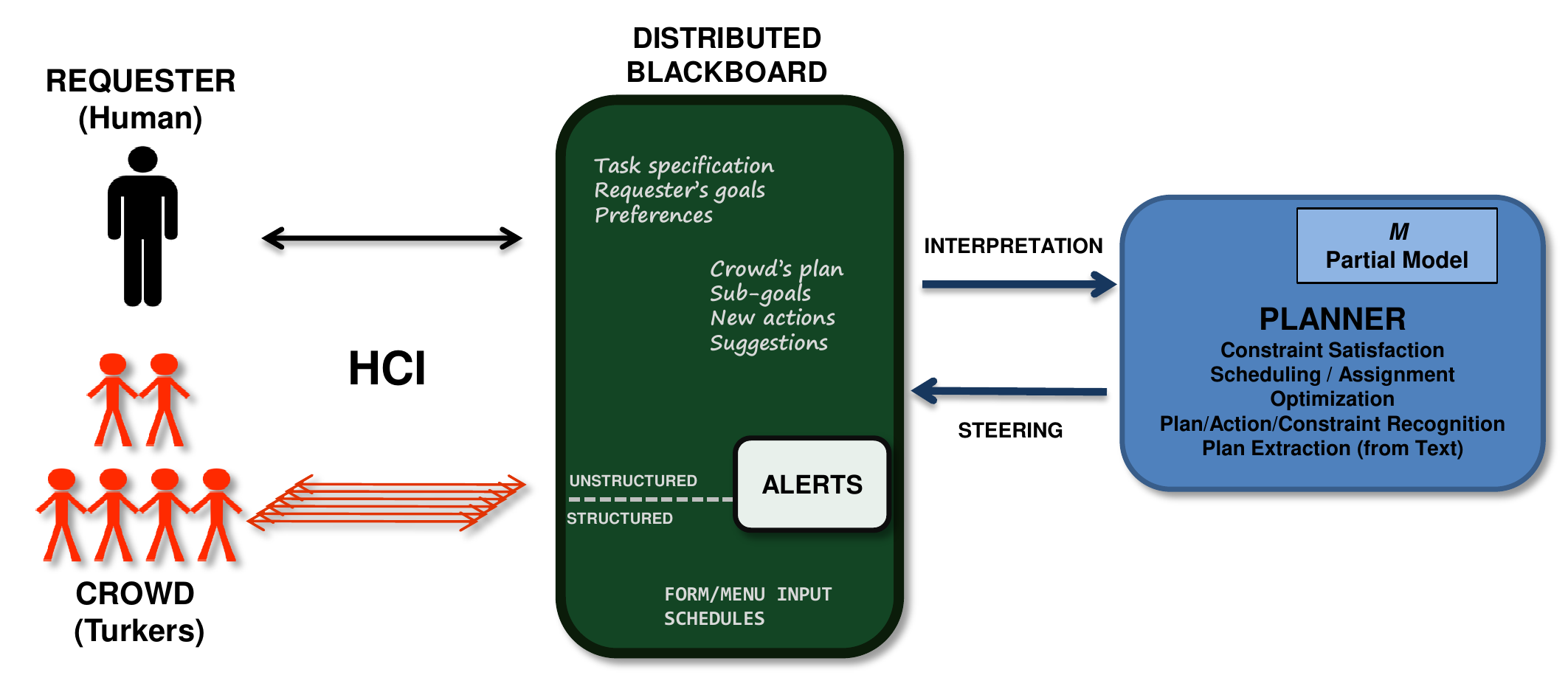}
    \caption{An architecture for planning in crowdsourced planning
      systems.
}
    \label{fig:crowdplan}
  \end{center}
\end{figure*}

The effectiveness of even primitive planning techniques begs the
obvious question: {\em is it possible to improve the effectiveness of
  crowdsourced planning even further by using more sophisticated
  automated planning technologies?}  It seems reasonable to expect
that a more sophisticated automated planner can do a much better job
of steering the crowd (much as human managers ``steer'' their
employees). It would also seem, at first blush, that importing
automated planning technology into crowdsourced planning scenarios
should be relatively straightforward. Indeed, work such as
\cite{law2011towards} and \cite{zhang2012human} are replete with
hopeful references to the automated planning literature.  There exists
a vibrant body of literature on automated plan generation, and
automated planners have long tolerated humans in their decision cycle
-- be it mixed initiative planning~\cite{ferguson1996trains} or
planning for
teaming~\cite{bagchi1996interactive,talamadupula2010tist}.
Nevertheless, the context of crowdsourced planning scenarios poses
some critical challenges in adapting planning technology.

In this paper, we aim to develop a general architecture for human
computation (crowdsourced) systems aimed at planning and scheduling
tasks, with a view to foreground the types of roles an automated
planner can play in such systems, and the challenges involved in
facilitating those roles. We shall see that the most critical
challenges include:

\begin{description}
\item[Interpretation] \blue{Need for interpreting the requester's
    goals as well as the crowd's plans from semi-structured or
    unstructured natural language input.}

\item[Steering with Incompleteness] Need for planning techniques that
  can get by with incomplete or incorrect models of both dynamics and
  preferences.


%

%

\end{description}
The {\em interpretation} challenge arises because human workers will
find it most flexible to exchange/refine plans expressed in a form as
close to natural language as possible, and automated planners
typically operate on more structured plans and actions.  The {\em
  steering} challenge is motivated by the fact that an automated
planner operating in a crowdsourced planning scenario cannot possibly
be expected to have a complete model of the domain or preferences; if
it does, then there is little need or justification for using human
workers!   Both these challenges
are further complicated by the fact that the (implicit) models used by
the human workers and the automated planner are very likely to differ
in many ways, making it hard for the planner to critique the plans
being developed by the human workers.  
We shall see that existing systems handle the incompleteness and
interpretation challenges in primitive ways, and subsequently discuss
ways in which these challenges can be handled in more effective
fashions.

The rest of the paper is organized as follows. We start by discussing
related work, and distinguish the focus of our work from them.  Next
we look at the problem of planning {\em for} crowdsourced planning in
more detail, and present a generalized architecture for this task.
Next, we consider the roles that an automated planner can play within
such an architecture, and discuss the challenges that need to be
tackled in order to facilitate those roles. We then describe a
spectrum of approaches for handling those challenges. Finally, we
characterize the existing crowdsourced planning systems in terms of
our architecture and challenges.
We hope that this work will spur directed research on the
challenges that we have identified.

\section{Related Work}
\label{sec:rel-work}

The role played by automated planning in crowdsourced planning
problems has interesting connections and contrasts to the role of
planners in mixed-initiative planning and planning for human-robot
teaming~\cite{talamadupula2010tist}. For example, in mixed-initiative
planning, the "interpretation" problem is punted away by expecting the
human in the loop to interact with the plan on the planner's terms;
this will certainly not work in crowdsourced planning. Further, in
mixed-initiative and human-robot teaming scenarios, the planner is
expected to have a complete model of the planning problem -- which is
rarely the case in crowdsourced planning. Instead, the planner must
deal with a model-lite~\cite{kambhampati2007mlp} spectrum, where
models may range from simple feasibility constraints, through
incomplete theories of the task domain and very rarely preferences
specified in a standardized format. Planning techniques that have so
far expected input in the form of PDDL (or some other standard) must
also change to take this model-lite spectrum into account.


A number of other implemented human computation systems that use
automated technology to assist with and improve the quality of tasks
other than planning are listed in~\cite{quinn2011human}'s wide-ranging
survey of the field. Our paper focuses solely on the crowd-planning
aspect, rather than the gamut of general human-computation tasks.

In this paper, we discuss how automated planning can and should help
crowd-sourced planning tasks. A related but different strand of work
is to use the planning technology in general crowd sourcing systems to
control the crowds (regardless of whether the task they are helping
with is a planning task or any other general computation task). An
example of this strand of research is the TurKontrol
project~\cite{dai2011artificial}, which is an end-to-end system that
dynamically optimizes live crowdsourcing tasks, and deals with the
problem of assigning human intelligence tasks (HITs) to both improving
the quality of a solution, as well as checking the current quality.
This work also concentrates on optimizing iterative, crowdsourced
workflows by learning the model parameters~\cite{weld2011human} from
real Mechanical Turk data, and modeling worker accuracy (for quality
improvement) and voting patterns (to check the quality of work done).
Such systems are independent of the actual task at hand -- whether
that be text improvement or human intelligence to produce plans -- and
focus more on worker-independent parameters to assign improvement and
voting jobs instead.

\section{Planning for Crowdsourced Planning}
\label{sec:planning}


The crowdsourced planning problem involves returning a plan as a
solution to a task, usually specified by a user or {\em requester}.
The requester provides a high-level description of the task -- most
often in natural language -- which is then forwarded to the crowd, or
{\em workers}. The workers can perform various roles, including
breaking down the high-level task description into more formal and
achievable sub-goals~\cite{law2011towards}, adding actions into the
plan that support those sub-goals~\cite{zhang2012human}, or propose
further refinements to the task (which can in turn be approved or
rejected by the requester, if they choose to remain part of the loop).
The {\em planner} is the automated component of the system, and it
performs various tasks ranging from constraint checking, to
optimization and scheduling, and plan recognition. The entire planning
process must itself be iterative~\cite{smith2012planning}, proceeding
in several rounds which serve to refine the goals, preferences and
constraints further until a satisfactory plan is found. A general
architecture for solving this crowdsourced planning problem is
depicted in Figure~\ref{fig:crowdplan}. 

\subsection{Motivating Example}
\label{subsec:example}

As a motivating (and running) example, we consider a problem that is
repeated around the world every year -- that of planning a {\em local
  tour}\footnote{Note that we only introduce, but do not formally
  define the Local Tour Planning problem in this paper.} for visitors,
families and students who are unfamiliar with a new college campus.
This problem can be easily generalized to any attraction that is {\em
  local} in nature; i.e., any attraction that is large enough to
warrant creating a customized plan, yet smaller than an entire city or
town, thus ruling out complex transportation planning problems. We
choose this problem because it offers a nice combination and trade-off
in terms of the causal complexity of the actions that any local tour
plan must contain, without transforming into a full-blown transit or
journey planning problem. Instead, we believe there are some rules
that are common to local tour planning problems across locations.
These common rules may be used as a guideline when creating the PDDL
model $M$ for this particular scenario.

In order to better illustrate the college tour problem, we present an
example request from a user, specified in natural language:
\textsl{\myquote{I want to take a tour of the State University
    campus. In particular, I want to see buildings that are relevant
    to a new undergraduate student, with an emphasis on the
    engineering departments. I also want to look at the various food
    options around the different places that I visit. I would like the
    tour to include some elements of the history and culture of the
    university. The tour should last about 3 hours, and I would like
    to finish at the parking spot where I started. It would also be
    nice to take a look at the football stadium.}}

\noindent In the above example, both the critical challenges -- {\em
  Interpretation}, and {\em Steering with Incompleteness} -- are
reflected to some degree. The user's preferences, expressed here in
the form of the vague goal as well as some partial preferences on what
they want to do, are incompletely specified. The dynamics of the
domain are incompletely known as well, both to the turkers that
receive this request from the user (the turkers may not know
everything about that specific campus), as well as to any automated
planner that must critique the turkers' plans.

The interpretation challenges come \blue{both in extracting goals from
  the requester, and identifying the plans being proposed/discussed by
  the crowd of turkers.}
It is not very likely that turkers will give
information about actions that satisfy the requester's preferences (a
tour) in a standardized form -- in fact, enforcing them to do such a
thing may lead to them leaving out important details. Instead, the
turkers must be encouraged to contribute as much information as
possible via a natural specification mode, e.g. free, unstructured
text. The automated planner's challenge is then to understand the plan
that the turkers are proposing from that free-form text. In order to
do this, the \blue{planner must be aware that its model of the campus
  may well be incomplete (and the human workers might have a better
  one)}.


Finally, once the crowd's plan is interpreted and structured knowledge
extracted from it, the automated planner must account for the fact that
the turkers created that plan using {\em their} own model of the
world -- a model that is decidedly closer to the requester's  as far
as preferences and goals are concerned, yet perhaps not as cognizant
of world constraints as the planner's model. Therefore the planner
must consider the gap between the two models when critiquing and
directing the further refinement of the crowd's plan. In the campus
tour planning example, it is easy to see how the planner may critique
a crowd plan that takes the requester to different corners of the
campus in the interests of time; yet the longer plan may actually be
fulfilling more of the requester's vague goal of wanting to ``see the
campus''. The optimization here is not merely on one factor (time
taken), but a complex combination of various factors (which humans are
better at even now than automated planners). 

In the rest of this paper, we will return time and again to this
running example, in order to illustrate more concretely some of the
points that we make.

\subsection{Roles in Crowdsourced Planning}

We will now get back to our 
general
architecture for  crowdsourced planning problem (shown in 
Figure~\ref{fig:crowdplan}), and take a more detailed look at the three major
roles in this problem:\footnote{We note  that we are considering a
  dedicated architecture for crowdsourced planning (much as systems
  such as DuoLingo consider dedicated architectures for specific tasks). An interesting
  future direction would be to consider how this architecture can be
  embedded into the popular crowdsourced platforms such as those
  provided by Amazon or Odesk.}

\subsubsection{Requester} 

The requester, or user, is at the head of this general system -- she
must specify the task at hand, as well as the desired goals; together,
we denote these as $G$. Additionally, the requester can specify
preferences on the form or contents of the plan that is eventually
returned as a solution. Usually, this specification is done in natural
language which can be understood easily by fellow (unskilled) humans,
but not as easily by machines and automated systems. The requester can
also (optionally) choose to observe the planning process as it
unfolds, and provide feedback in order to steer  it in the right
direction. An example of a specification from a requester is provided
in Section~\ref{subsec:example}.

\subsubsection{Workers (Crowd)}

The task specification $G$ is then passed on to the crowd of workers.
The workers' job consists of two main tasks: 

\begin{enumerate}

\item Break down $G$ into a set of machine-readable sub-goals
  $S_G$. An example of this can be seen in
  Section~\ref{subsec:example}, where the workers must take the
  requester's vaguely specified goal of wanting a tour of campus into
  smaller sub-goals such as visiting specific buildings and conducting
  various activities.

\item Help generate a ``crowd'' plan $P_c$ aimed at solving $G$,
  either by suggesting a brand new one or by modifying the existing
  plan based on critiques. This part consists of the workers
  suggesting actions that will fulfill the sub-goals specified by them
  in the previous stage.

\end{enumerate}

The crowd may also provide preferences on the task, either explicitly
or by evaluating plans that were generated in previous
iterations\footnote{When the requester and crowd's preferences clash,
  it is usually the case that the requester wins -- since they are
  paying for the crowdsourced planning process.} -- for example, there
may be many different tours of a campus that visit different buildings
via different routes, and it is up to the workers to choose the one
that they think best satisfies the requester's specified preferences.
Note that $P_c$ may be generated in natural language, and does not
need to have a formal structure, since it involves multiple workers
collaborating and natural language is the lowest common denominator
when it comes to workers of various skill-levels collaborating.

\subsubsection{Planner}

The planning module, or the automated component of the system, can
provide varying levels of support. It  accepts
both $S_G$ and $P_C$ as input from the workers. This module analyzes
the current plan generated by the crowd, as well as the sub-goals, and
determines constraint violations according to the model $M$ of the
task that it has (see Section~\ref{sec:related}). The planner's job is
to steer the crowd towards more effective plan generation. 


\noindent However, these three roles need a common space in which to
interact and exchange information. This is achieved through a common
interactive space -- the {\em Distributed Blackboard} (DBb) -- as
shown in Figure~\ref{fig:crowdplan}. The DBb acts as a collaborative
space where all the information related to the task as well as the
plan that is currently being generated to solve it is stored, and
exchanged between the various components of the system.

In contrast to the workers, the planner cannot hope for very complex,
task-specific models, mostly due to the difficulty of creating such
models. Instead, a planner's strong-suit is to automate and speed-up
the checking of plans against whatever knowledge it {\em does}
have. With regard to this, $M_P$ can be considered shallow with
respect to preferences, but may range the spectrum from shallow to
deep where domain physics and constraints are concerned. 
\\

The planning process itself continues until one of the following
conditions (or a combination thereof) is satisfied:

\begin{itemize}

\item The crowd plan $P_c$ reaches some satisfactory threshold and the
  requester's original goal $G$ is fulfilled by it; this is a
  subjective measure and is usually determined with the intervention
  of the requester.

\item There are no more outstanding alerts, and all the sub-goals in
  $S_G$ are supported.

\end{itemize}

\noindent The various interactions among the requester, workers and
the planner produce several problems of interest to automated
planning. In the following, we categorize these problems into two main
categories, and we then describe the constituents of these
categories -- problems that should be familiar to the automated
planning community, yet are  missing from current crowdsourced
planning systems.

\nocite{gupta2006creating}
\nocite{kokkalis2012automatically}
\nocite{mao2012turkserver}
\nocite{zhang2013automated}
\nocite{kittur2013future}
\nocite{lasecki2012speaking}
\nocite{chilton2013cascade}
\nocite{hung2013leveraging}
\nocite{branavan2009learning}

\section{Planning Challenges}

From the architecture described in Figure~\ref{fig:crowdplan}, it is
fairly obvious that a planner (automated system) would interact with
the rest of the system to perform one of two tasks: (1) {\bf
  interpretation} and (2) {\bf steering}. 
These two tasks define the planner's role in the entire
process. Interpretation is required for the planner to inform itself
about what the crowd {\em is} doing; steering is required for the
planner to tell the crowd what they {\em should} be doing. Here we
take a deeper look at these two modes.

\subsection{Interpretation of the Crowd's Evolving Plan}

The planner must interpret the information that comes from the
requester, and from the crowd (workers), in order to act on that
information. There are two ways in which the planner can ensure that
it is able to understand that information: 

\subsubsection{Force Structure}

The system can enforce a pre-determined structure on the input from
both the requester, and the crowd. This can by itself be seen as part
of the model $M_p$, since the planner has a clear idea about what kind
of information can be expected through what channels. For example, in
a travel-planning application, the requester can be given a dynamic
form to fill out, instead of a box for free-form text. This instantly
imposes structure on the information provided, and makes it easier for
the planner to separate various fields. Similarly, when the workers
are creating plans, the system can impose a flow on that process. Such
structured data can include -- apart from just the dynamics of the
domain in question -- information about the requester's preferences as
well. The obvious disadvantage is that it reduces flexibility for the
human workers.

An interesting research challenge here is to develop
interfaces that will incentivize human workers to provide more
structured 
information about their plans--including temporal, causal and teleological
dependencies. 
\blue{In the campus tour example, we might force the requester to
  number his/her goals, and force the turkers to explicitly state
  which goals their proposed plan aims to handle
  (c.f. \cite{zhang2012human}). We might also force the turkers to add
 other  structured attributes to their plans such as the amount of time that
  is expected to be taken by the plan.}


\subsubsection{Extract Structure}

The planner can also {\em extract} structure from the text input of
the \blue{(human) requester as well as} human workers, in order to
determine the current state of the crowd-planning process.  The
specific extraction method used may vary from methods that extract
from plain text and impose structure~\cite{ling2010temporal}, to plan
extraction which tries to obtain a structured plan from unstructured
text. Although this problem has connections to plan recognition
\cite{kautz1986generalized,geffner-ramirez}, it is significantly
harder as it needs to recognize plans not from actions, but rather
textual descriptions.  Thus it can involve first recognizing actions
and action order from text, and then recognizing plans in terms of
those actions.


\blue{Consider an example from the campus tour problem -- turker input
  such as \textsl{\myquote{`I suggest you go to the student union at
      lunch time as there are many entertainment shows at that time''
    }}

  \noindent would have to be interpreted in terms of the action (of
  going to the student union). Subsequent to this, the planner must
  also identify additional information related to that action, such as
  the time when it is to execute (lunch time) and possible goals that
  it satisfies (entertainment).  }



Unlike traditional plan recognition that starts from observed plan
traces in terms of actions or actions and states, the interpretation
involves first extracting the plan traces.  The extracted traces are
likely to be noisy (e.g.~\cite{hankz-nips-2012}) complicating plan
recognition. An even more challenging obstacle is the impedance
mismatch between the (implicit) planning models used by the human
workers and that available to the planner.

\subsection{Steering the Crowd's Plan}

After determining what is going on in the planning process, the
planner can steer the workers by offering helpful suggestions, alerts
and perhaps even its own plan. 
In some ways, the role of the planner in this scenario is akin to that
of a human manager who effectively oversees, shepherds,  and steers employees
without necessarily knowing the full details of what the employees are
doing \cite{march1958organizations,davis1997toward}. 
There are two main kinds of feedback an automated planner can provide
to the human workers:


\subsubsection{Problem Identification}

Even with a very simple model, the planner can be used as a basic
automated constraint and arithmetic checker. For example, in producing
crowdsourced plans for travel planning, if the only things known as
part of the automated planner's model are the maximum distance that
the requester is willing to walk, and the number of transfers they
want to make -- as is often the case with transit directions in online
maps -- then the automated planner is restricted to just enforcing
those constraints on any plan that the crowd comes up with. A planner
with a slightly more complex model can try to apply the plan
recognition methods described previously, in order to generate alerts
for the crowd in terms of sub-goals or actions that are currently
unsupported.

\subsubsection{Constructive Critiques}

Once the planner has some knowledge about the plan that the workers
are trying to propose (using the recognition methods described above),
it can also try to actively help the creation and refinement of that
plan by offering suggestions as part of the alerts. These suggestions
can vary depending on the depth of the planner's model. They can range
from simple notifications of constraint violations, as outlined
previously; through plan critiques (such as suggestions on the order
of actions in the plan and even what actions must be present);
suggesting new plans or plan fragments
because they satisfy the requester's
stated preferences or constraints better; and suggesting new ways of
decomposing the problem~\cite{shop}.


As far as the actual generation of suggestions goes, given an
incomplete model $M$, a simple regression approach can be adopted that
tries to match the sub-goals in the scenario with the actions that
support them (those actions having been extracted previously from
text). This approach can be augmented further with the introduction of
plan, goal and intent recognition methods. The planning system can try
to guess which of the sub-goals are currently being supported by the
actions in the crowd's plans, and expand that particular path further
in order to generate alerts that are more specific to the current plan
under consideration. Additionally, there are connections to other,
established planning and scheduling problems:

 \subsubsection{Model Evolution}

 Given that the crowd's plan $P_c$ -- being realized as free-form
 natural language expressed as text -- may contain actions not present
 in $M$, the planner may also ask the human planners to (i)
 ``explain'' the role of those actions with respect to subsequent
 actions, or (ii) confirm the applicability of those unknown actions
 with respect to preceding actions. These alerts may help the planner
 update the model $M$ with preconditions and effects of the new
 actions.

 \subsubsection{Preference Handling \& Elicitation} 

 Approaches can range from already implemented methods, like
 generating a diverse set of plans for the crowd or the requester to
 pick from~\cite{nguyen2012generating} (implicit preference
 elicitation), to making the crowd explicitly enumerate the
 preferences that the requester might hold (which may also have been
 specified via natural language on the DBb).

 \subsubsection{Scheduling \& Optimization}

 In certain cases, the crowd produces suggestions for actions that can
 be used to create a plan for the requester's task. However, those
 actions still need to be scheduled to create the plan $P_c$. The
 automated system can be used to perform this scheduling -- in certain
 cases, if the model is detailed enough, the system can even be used
 to perform optimization to produce the best plan from the suggestions
 mooted by the crowd.

 \subsubsection{Differences from traditional plan
   synthesis/critiquing}

 While the task of plan steering has several similarities to the
 traditional plan synthesis and plan critiquing \cite{dana-textbook},
 it differs in significant ways because of the incompleteness of the
 domain models and requester preferences available to the planner. The
 model incompleteness precludes the traditional techniques that view
 planning as producing a provably correct course of action.  The
 incompleteness of the model, as well as the attendant impedance
 mismatch between the planner's and human workers' models also makes
 the plan critiquing harder. What may be seen as a wrong or suboptimal
 plan given the planner's incomplete model of the domain may well be a
 desirable one from the requester's point of view. This difficulty is
 ameliorated in part by ensuring that the planner only provides
 non-binding alerts/advice to the human workers.

 Rather than traditional planning models, we believe ``model-lite
 planning,'' as envisioned in
 \cite{kambhampati2007mlp,nguyen2009planning,james-allen-plow} may be
 more appropriate for crowdsourced planning scenarios. In particular,
 \cite{kambhampati2007mlp} categorizes planning with incomplete models
 into two cases, shallow model case and approximate model case,
 depending on the degree of incompleteness of the domain model. We
 believe that this distinction is relevant for crowdsourced planning
 scenarios too.  In particular, {\em approximate} domain models are
 those that are almost complete, but have some missing details.
 Examples of missing details could include missing preconditions and
 effects of actions (c.f. \cite{nguyen2009planning,bryce-1,bryce-2}),
 or cost models. We would like to be able to use approximate models to
 support plan creation as well as plan critiquing.  {\em Shallow}
 domain models, in contrast, are those that aim to provide knowledge
 to mostly support critiquing, rather than creation of plans. Examples
 of shallow models include I/O type specifications, task dependency
 knowledge, or databases of past plans (aka case-bases), or even
 low-level constraints (such as temporal deadlines) etc.  Typically,
 these models are not generative, and do not involve
 precondition-effect style characterization of the actions. They are
 useful mostly for critiquing the plans (c.f.
 \cite{james-allen-plow,woogle}). 
 (It is of course possible to have domain models that are shallow in
 some aspects and approximate in other.)

 Another aspect of incomplete models is that the planning has an
 incentive to improve the completeness of its model over time. It will
 be interesting to see if the existing work on learning planning
 models (c.f. \cite{journal/aij/Yang07,blythe-learn}) can be adapted
 to allow learning from observing the crowd's plans.

 \section{Classifying Existing Crowdsourced Planning Systems}
\label{sec:related}

In the previous section, we saw that both the challenge of interpreting the crowd's
plan and the challenge of steering it can have primitive solutions
(e.g. force structure and critique the plan in terms of lower level
consistency checks), and more ambitious solutions (e.g. interpret
structure by extracting actions and plans from text, and evaluate the
extracted plan in terms of the planning model to provide constructive
extensions or alternatives for the crowd's consideration). We shall
see in Section~\ref{sec:related} that most existing work uses the
primitive solutions for interpretation and steering. Their success
argues for exploration of the more ambitious solutions to these
problems.

A few systems have attempted to solve some version of the
crowdsourced planning problem.  All of these systems can be seen as
special cases of the general architecture shown in
Figure~\ref{fig:crowdplan}. In the following, we describe 
approaches that rely on automated systems in order to improve the
synthesis of crowd-plans, or the quality of those plans.



Mobi~\cite{zhang2012human} takes a planning mission that consists of
both preferences and constraints as input from a requester, and
generates a plan or itinerary by allowing workers in the crowd to plan
in a shared manner.  Constraints are limited to two types:
qualitative, which are high-level and specified in natural language
(e.g. what the user hopes to accomplish with the trip); and
quantitative, which are specified over arbitrary categories that may
be created by requesters (e.g.  ``cool artsy things'').  Constraints
may be specified either over the amount of time to be spent on
activities in each category, or on the number of such activities.
Taken together, these can be seen as Mobi's primitive model (``shallow
model'' in the terminology of the previous section), which is enforced
by a simple automated constraint checker.  Zhang et al. show in two
experiments that: (i) for the same amount of money spent on human
workers, a system with automated alerts tends to come up with higher
quality plans; and (ii) the automated alerts tend to spur the plan
towards breaching a set plan quality threshold in far fewer steps than
a system without them.

Law and Zhang \cite{law2011towards} 
introduce CrowdPlan, a collaborative planning algorithm that takes as
input a high-level mission from the user (such as ``I want to live a
healthier life'') and provides web-based resources for accomplishing
that mission. To facilitate this, CrowdPlan uses human workers to
decompose the high-level mission into a variety of goals 
(such as ``stop smoking'', ``eat healthier food''). Although the decomposition process
has similarities to HTN planning \cite{shop}, CrowdPlan itself doesn't
have any automated planning component overseeing the human workers. 

On the other hand, there are systems like
CrowdPlanr~\cite{lotosh2013crowdplanr} that focus more on sequencing
the steps in a plan once the actions themselves have been
selected. CrowdPlanr takes a given set of actions -- for example, in a
travel planning scenario, the cities in Italy that one could visit
starting from Rome -- and determines the least number of questions
(and what those questions are) to ask the crowd of workers to achieve
a plan of acceptable quality. The requester initially specifies the
constraints associated with the task in free-form text (for e.g.,
``the trip must last 2 weeks'') and it is assumed that the crowd will
take these into consideration when answering the questions posed by
the system. It is important to note the departure from the two
previous systems we have looked at -- in this case, the {\em model}
consists of both the constraints specified by the requester as well as
the crowd's knowledge. It is quite likely that the quality of plans
produced by such a system would be sensitive to the familiarity of the
workers with the task at hand.

The Cobi~\cite{cobi} system employs the same basic idea -- that the
{\em crowd} that is assisting with the planning already has a built-in
model of preferences and constraints. Cobi seeks to
``communitysource'' the scheduling of a large-scale conference (CHI
2013) by taking input from organizers, as well as authors and
attendees in order to come up with a schedule (plan) of good quality,
that violates a fewer number of constraints while being feasible. The
automated system performs four overall tasks: (i) clustering papers by
topic into either sessions or affinity groups larger than sessions;
(ii) preference collection, both hard and soft; (iii) scheduling of
rooms and time slots of sessions; and (iv) assignment of session
chairs based on the best session matches to a person's expertise. The
collection of these constraints, and the grouping of papers into areas
of expertise for the clustering, may be seen as Cobi's model. The
automated system thus uses this model in order to resolve as many of
the constraints as possible, and come up with a conference schedule
which is both satisfactory, and more importantly transparently
collaborative.


%

The system proposed (but not yet implemented)
by~\cite{lasecki2012real} is the closest in spirit and idea to
applying automated planning methods on a distributed interaction
platform to aid crowdsourced planning. That system allows workers to
decompose tasks and interact by posting constraints and further
sub-tasks to the problem of planning a trip.  The task is specified by
the requester in free-form text, using natural language; some
constraints may be specified as part of the task (for e.g., date and
time constraints, cost ceilings etc.).  Workers are provided up-front
with four decomposable sub-tasks -- a primitive model of sorts -- into
which further sub-tasks may be added.  Additionally, workers are
provided with a text box into which they can type suggestions for new
tasks and constraints. The automated system enables collaboration
amongst the workers and the requester, and also extracts additional
information from their input (restricted to a structured form) in
order to update the model.

\section{Conclusion}
In this paper, we took a first step towards investigating the
opportunities and challenges for automated planning technology in
crowdsourced planning scenarios. We identified several roles an
automated planner can play in steering the human workers in producing
effective plans. We then identified two important challenges in
adapting automated planning technology to such scenarios: {\em
  interpreting} the requester inputs as well as human worker
plans--often expressed in natural language, and critiquing these plans
in the presence of {\em incompleteness} of requester preferences as
well as the planner domain model. We discussed several ways in which
these challenges can be tackled, and also characterized the specific
(if primitive) choices made by the existing crowdsourced planning
systems in handling these challenges.
We hope that this work will spur directed research on the
challenges that we have identified. 

\bibliographystyle{aaai}
\bibliography{hcomp}

\end{document}